\documentclass[letterpaper]{article} 
\usepackage{aaai24}  
\usepackage{times}  
\usepackage{helvet}  
\usepackage{courier}  
\usepackage[hyphens]{url}  
\usepackage{graphicx} 
\usepackage{natbib}  
\usepackage{caption} 
\usepackage{algorithm}
\usepackage{algpseudocode}

\usepackage[utf8]{inputenc} 
\usepackage{booktabs}       
\usepackage{amsfonts}       
\usepackage{amsmath}
\usepackage{nicefrac}       
\usepackage{microtype}      
\usepackage{threeparttable}
\usepackage{subcaption}
\usepackage{multirow}

\usepackage{newfloat}
\usepackage{listings}
\DeclareCaptionStyle{ruled}{labelfont=normalfont,labelsep=colon,strut=off} 
\usepackage[dvipsnames]{xcolor}

\title{TE2Rules: Explaining Tree Ensembles using Rules}
\author {
    G Roshan Lal, \hspace{2mm}
    Xiaotong (Elaine) Chen, \hspace{2mm}
    Varun Mithal
}
\affiliations {
    LinkedIn AI\\
    rlal@linkedin.com, \hspace{2mm} elachen@linkedin.com, \hspace{2mm} vamithal@linkedin.com\\
}

\nocopyright

\usepackage{bibentry}

\begin{document}

\maketitle

\begin{abstract}

Tree Ensemble (TE) models, such as Gradient Boosted Trees, often achieve optimal performance on tabular datasets, yet their lack of transparency poses challenges for comprehending their decision logic. This paper introduces TE2Rules (Tree Ensemble to Rules), a novel approach for explaining binary classification tree ensemble models through a list of rules, particularly focusing on explaining the minority class. Many state-of-the-art explainers struggle with minority class explanations, making TE2Rules valuable in such cases. The rules generated by TE2Rules closely approximate the original model, ensuring high fidelity, providing an accurate and interpretable means to understand decision-making. Experimental results demonstrate that TE2Rules scales effectively to tree ensembles with hundreds of trees, achieving higher fidelity within runtimes comparable to baselines. TE2Rules allows for a trade-off between runtime and fidelity, enhancing its practical applicability. The implementation is available here:\url{https://github.com/linkedin/TE2Rules}.

\end{abstract}

\section{Introduction}
\label{sec:introduction}

In recent years, many decision support systems have been constructed as black box models using machine learning such as Tree Ensembles (TE) and Deep Neural Networks. Lack of understanding of the internal logic of decision systems constitutes both a practical and an ethical issue, especially for critical tasks that directly affect the lives of people like health care, credit approval, criminal justice etc. In such use cases, there is a possibility of making wrong decisions, learned from spurious correlations in the training data. The cost of making wrong decisions in these domains is very high. Hence, having some explanations (like what part of the input is the model focusing on, or under what conditions satisfied by the input, does the model behave similarly) is important for building trust on these decision systems. Moreover, recent legal regulations like General Data Protection Regulation (GDPR, 2018) enables all individuals to obtain ``meaningful explanations of the logic involved'' when automated decision making takes place.

In this work, we focus our attention on explaining tree ensemble (TE) models which are popular in many use cases involving tabular data \cite{grinsztajn2022treebased, qin2021are}. Our focus is exclusively on binary classification models, as they are prevalent in many critical decision-making systems such as disease diagnosis and spam detection. In the realm of binary classification, providing explanations for one class can be more crucial than the other. Consider healthcare or fraud detection, where explaining why a model identifies a data point as positive (e.g., detecting a tumor or predicting a scammer) is vital. Interestingly, the positive class often represents the minority class in the training/test data.

A popular method to explain any model is to learn an interpretable surrogate model that closely approximates the original model. The accuracy of the surrogate model with respect to the original model is called \emph{fidelity}. A good explainer needs to have high fidelity on test data. Additionally, for effective explanation of the minority class, the explainer must demonstrate good fidelity specifically on the part of the test data where the model predicts the minority class. Therefore, the explainer needs to maintain high fidelity overall and also high fidelity on the minority class predictions. However, many state of art rule-based explainers excel in overall fidelity but struggle to explain the minority class accurately. This is problematic, especially when the minority class corresponds to the positive class. This limitation hinders their utility in explaining predictions for the important positive class. 

In this work, we introduce a novel algorithm TE2Rules (Tree Ensemble to Rules) designed to mine rules only for the (minority) positive class. Each individual rule takes the form "If feature$_i > t_i$ and feature$_j \leq t_j$ \ldots and feature$_k \leq t_k$, then model prediction = 1," where the label 1 signifies the positive class. The rules mined by TE2Rules are short with only a few features and each rule has a high precision, i.e, of all the data points that satisfy the rule, a high fraction of them (default of 95\%) get positive class prediction by the model. Besides high precision, each rule has a decent coverage on positives. We post process these rules by selecting a small number of rules that cover most of the positives in the data. This small collection of rules can explain the model at a global level with high overall fidelity as well as high fidelity on positive class predictions. TE2Rules can mine these rules in a runtime that is comparable to other state of the art model explainers making the algorithm scale to tree ensembles with hundreds of trees.

TE2Rules achieves it capabilities by leveraging the Apriori Algorithm. The Apriori algorithm is a data mining algorithm that identifies items that are frequently found together in a collection of itemsets. In the context of TE2Rules, positive class predictions are explained by identifying sets of internal tree nodes that are commonly active with positive class predictions, but not with negative class predictions. These sets of tree nodes are then converted into rules using the necessary conditions that an input data point must satisfy to traverse that particular combination of tree nodes within the tree ensemble.

In this work, we show that 1) many existing state of the art rule based explainers for tree ensemble (TE) models have poor fidelity on positives (the minority class) though they may have good overall fidelity. 2) To solve this problem, we propose a novel method, TE2Rules (Tree Ensemble to Rules), that can generate rules corresponding to a single class of interest (say positive class) by merging decision paths from multiple trees in the tree ensemble. Since all the rules are mined for a single class, there is no conflict among the labels predicted by different rules. 3) We show that the resulting rules have high overall fidelity and high fidelity on positives (the class of interest) even if the positives happen to be a minority class in the dataset. TE2Rules can achieve such high performance at comparable number of rules relative to existing baselines, at the cost of slightly higher run time. 4) By stopping the algorithm in the early stages, we can tradeoff fidelity with runtime.

\section{Related Work}
\label{sec:related}

\begin{figure*}
\begin{minipage}{0.27\linewidth}
    \centering
    \begin{tabular}{ | c|c | } 
    \hline
    \textbf{id} & \textbf{itemsets} \\ 
    \hline
    0 & \{bread, milk\} \\ 
    \hline
    1 & \{jam, milk\} \\ 
    \hline
    2 & \{jam, butter\} \\ 
    \hline
    3 & \{bread, jam, milk\} \\ 
    \hline
    4 & \{bread, jam, milk\} \\ 
    \hline
    5 & \{bread, jam, butter\} \\ 
    \hline
    6 & \{bread, jam, eggs\} \\ 
    \hline
    \end{tabular}
    \caption{Dataset}
    \label{tab: itemset}
\end{minipage}%
\vline\vline\vline
\begin{minipage}{0.24\linewidth}
    \centering
    \begin{tabular}{ | c|c | } 
    \hline
    \textbf{itemsets} & \textbf{count}\\ 
    \hline
    \textcolor{Green}{\{jam\}} & \textcolor{Green}{6}\\ 
    \hline
    \textcolor{Green}{\{bread\}} & \textcolor{Green}{5}\\ 
    \hline
    \textcolor{Green}{\{milk\}} & \textcolor{Green}{4}\\ 
    \hline
    \textcolor{Green}{\{butter\}} & \textcolor{Green}{2}\\ 
    \hline
    \textcolor{Red}{\{eggs\}} & \textcolor{Red}{1}\\ 
    \hline
    \end{tabular}
    \caption{Apriori Stage 1}
    \label{tab: stage1}
\end{minipage}%
\begin{minipage}{0.23\linewidth}
    \centering
    \begin{tabular}{ | c|c | } 
    \hline
    \textbf{itemsets} & \textbf{count}\\ 
    \hline
    \textcolor{Green}{\{jam, bread \}} & \textcolor{Green}{4}\\ 
    \hline
    \textcolor{Green}{\{jam, milk\}} & \textcolor{Green}{3}\\ 
    \hline
    \textcolor{Green}{\{jam, butter\}} & \textcolor{Green}{2}\\ 
    \hline
    \textcolor{Green}{\{bread, milk\}} & \textcolor{Green}{3}\\ 
    \hline
    \textcolor{Red}{\{bread, butter\}} & \textcolor{Red}{1}\\ 
    \hline
    \textcolor{Red}{\{milk, butter\}} & \textcolor{Red}{0}\\
    \hline
    \end{tabular}
    \caption{Apriori Stage 2}
    \label{tab: stage2}
\end{minipage}%
\begin{minipage}{0.28\linewidth}
    \centering
    \begin{tabular}{ | c|c | } 
    \hline
    \textbf{itemsets} & \textbf{count}\\ 
    \hline
    \textcolor{Green}{\{jam, bread, milk\}} & \textcolor{Green}{2}\\ 
    \hline
    \end{tabular}
    \caption{Apriori Stage 3}
    \label{tab: stage3}    
\end{minipage}
\vspace{-4mm}
\end{figure*}

In the past, several methods have been proposed to explain a tree ensemble model using rules or decision trees that can be better understood by a human. Some of these approaches work at a global level by approximating the tree ensemble with a set of rules or decision tree. While some other approaches work at a local (instance-level) by finding a rule or a decision tree that best explains the decisions of the tree ensemble for data points sampled from the neighborhood of that instance.

\textbf{Rule-based explainers}: inTrees\cite{deng2019interpreting} generate rules from decisions made from individual trees in the tree ensemble and selects high precision rules among them. Another closely related method, ruleFit \cite{rulefit2008} runs a sparse linear regression on rules generated from individual trees to select the most important rules. However, a sparse ensemble of rules is not as interpretable as a list of if-then rules. Both inTrees and ruleFit generate rules from nodes of individual trees and do not consider node combinations from multiple trees. Hence, their search space of rules is limited to rules from individual trees, resulting in low fidelity. deFragTrees \cite{defragtrees2018} identifies fragmented regions in the input space defined by the splits made by the tree ensemble and tries to simplify them into a short set of rules that are almost equivalent to the tree ensemble using bayesian inference. deFragTrees works on simplifying rules obtained from all possible node combinations from multiple trees in the ensemble and can achieve higher fidelity than methods like inTrees. However, most rule-based explainers are not targeted to explain any one single class. They often end up mining a lot of rules for the majority class and miss out on the minority class. In such cases, they are not very effective in explaining the minority class prediction.

\textbf{Tree-based explainers}: BATrees \cite{vidal2020bornagain}, uses the tree ensemble to generate more labeled data points that can be used to fit a decision tree on the data. However, trying to obtain a single tree decision tree to represent a tree ensemble can result in a very deep tree, making it harder to interpret. Node Harvest \cite{node_harvest2010} selects a few nodes from the shallow parts of the trees in the ensemble and creates an ensemble of shallow trees. The simplified model is easier to interpret than the original model. But, it is still not as easy to interpret as a decision tree or rule list.

\textbf{Local instance-level explainers}: Some explanation methods are model-agnostic and can be applied to models beyond tree ensembles. Anchors \cite{anchors} finds high precision, if-then rules satisfied by the instance, using multi-armed bandit and beam search algorithms. LoRE (Local Rule-based Explanations) \cite{lore} constructs interpretable models (decision trees) based on local samples. For each input data point, rules are generated that are locally accurate. A local search is conducted for every incoming data point to be explained. Hence, these methods can be prohibitively expensive if the objective is to explain a very large set of data points for a single model. 

\textbf{Interpretable models}: Instead of explaining existing machine learning models, an alternative approach is to directly learn interpretable models from data. Some methods, like Falling Rule List \cite{frl2015}, Interpretable Decision Sets \cite{lakkaraju2016interpretable}, and Bayesian Rule List \cite{brl2017, wang2017bayesian}, learn rule lists directly from data using association rule mining algorithms such as Apriori \cite{apriori1994} or its variants like FP-Growth \cite{fpgrowth2005}. 

The Apriori Algorithm is a data mining algorithm designed to identify subsets of items that frequently co-occur in a collection of itemsets. Apriori is particularly valuable in analyzing e-commerce transactions. For instance, in a database of customer transactions where each transaction includes a set of items bought together (like $\{$milk, bread, jam$\}$, $\{$eggs, bread, jam$\}$, etc.), Apriori can identify frequently co-occurring subsets of items (such as $\{$bread, jam$\}$). 

Some other approaches like SkopeRules \cite{skoperules2021} learns rule lists by first fitting a tree ensemble model on the data and then extracting rules from it. Unlike the methods that explain a trained tree ensemble model, SkopeRules learns a tree ensemble model internally and doesn't explain an existing trained tree ensemble model.

In this work, our focus is on explainers that take a tree ensemble model and some data as input, producing rules that explain model prediction, especially for the minority (positive) class. The explainer operates globally, producing different possible rules under which the model gives positive predictions. For this purpose, we utilize the Apriori Algorithm to identify frequently co-occurring decisions made within the nodes of the tree ensemble that result in positive class predictions. While the Apriori Algorithm has been employed in recent times to learn interpretable models directly from data, our work represents the first application of association rule mining for explaining a trained tree ensemble model. We choose inTrees and deFragTrees as our baselines since they are also global explainers that generate a rule list from the tree ensemble model and a slice of training data.

\section{Method}
\label{sec:method}

\textbf{Apriori Algorithm}: Before describing TE2Rules, it is important to understand the Apriori Algorithm. Apriori is a data mining algorithm designed to identify subsets of items that frequently appear together in a collection of itemsets. It uses 2 user-defined parameters: $m$ and $k$. Apriori looks for subsets containing $1, 2, 3,$ upto $k$ items, where the items within each subset occur together more than $m$ times in the overall collection of itemsets. For example, Table \ref{tab: itemset} shows a dataset with a collection of 7 itemsets. The itemset \{bread, jam\} appears as a subset in 4 different itemsets out of 7 in the dataset. Apriori tries to find such frequently occurring subsets. Apriori algorithm runs in stages. In the $k$-th stage, it tries to find subsets of size $k$ that occur more than $m$ times. Here's a brief description of how Apriori works.

In stage 1, Apriori examines individual items or $1$-item sets that have a frequency more than $m$ in the collection. Such, itemsets are considered to be frequent enough in stage 1.

In stage $k'$ (with $1 < k' \leq k$), it identifies itemsets of size $k'$ that occur more than $m$ times in the dataset. To generate candidate itemsets for the current stage $k'$, the Apriori algorithm looks at pairs of itemsets of size $(k'-1)$ from the previous stage. Combining such a pair would result in an itemset of size $2(k'-1)$. However, Apriori only considers pairs of itemsets that have $(k'-2)$ items in common. Combining such a pair results in itemset with $2(k'-1) - (k'-2) = k'$ items. 

The algorithm leverages the ``Anti-monotone property," which states that if an itemset is frequent, then all of its subsets must also be frequent. For each candidate itemset generated above, Apriori checks if all its subsets of size $(k'-1)$ have been found to be frequent in the previous stage.

For a candidate itemset that has passed the test of ``Anti-monotone property," Apriori counts its number of occurences in the dataset and keeps only those candidate itemsets that occur more than $m$ times in the dataset. These itemsets are considered to be frequent enough in stage $k'$.

\textbf{Example}: Let's apply Apriori to discover subsets of items that appear more than $m$ = 1 times in Table \ref{tab: itemset} by running Apriori till $k=3$ stages.

\textbf{Stage 1}: In stage 1, Apriori considers individual items or sets of $1$ item, that have a frequency exceeding $m$ in the collection. The counts of these items in the collection are presented in Table \ref{tab: stage1}. Among these items, \{jam\}, \{bread\}, \{milk\}, \{butter\} are identified as frequent since they occur more than $m$ = 1 times in the collection.

\textbf{Stage 2}: In stage 2, Apriori combines pairs of itemsets found to be frequent in stage 1 to form itemsets of 2 items. This ensures that itemsets containing \{eggs\} are excluded from exploration in stage 2, since the itemset \{eggs\} was not frequent enough in stage 1. The counts of these 2-item itemsets formed in this manner are presented in Table \ref{tab: stage2}. Among these items, \{jam, bread\}, \{jam, milk\}, \{bread, milk\}, \{jam, butter\} occur more than $m$ times in the collection.

\textbf{Stage 3}: In stage 3, Apriori combines pairs of itemsets that were identified as frequent in stage 2. If a pair of itemsets shares one item in common, Apriori merges them to create a new itemset with three items. For instance, it combines \{jam, bread\} and \{jam, milk\} to produce \{jam, bread, milk\}, and similarly for other combinations.

After generating these 3-item itemsets, Apriori verifies that all 2-item subsets derived from each itemset was found to be frequent enough in stage 2. For example, the itemset \{jam, bread, butter\} is formed from \{jam, bread\} and \{jam, butter\}. But, the itemset \{jam, bread, butter\} also contains the subset \{bread, butter\}, which was not frequent in stage 2. Thus, Apriori removes this 3-item itemset. Following this approach, only the 1 itemset remains: {jam, bread, milk} with all its subsets already identified as frequent in stage 2. This itemset also meets the frequency threshold by appearing more than $m$ times in the collection (as illustrated in Table \ref{tab: stage3}).

Thus, the itemsets \{jam\}, \{bread\}, \{milk\}, \{butter\}, \{jam, bread\}, \{jam, milk\}, \{bread, milk\}, \{jam, butter\}, \{jam, bread, milk\} occur frequently (more than $m$ times) in the collection.

\begin{figure*}[ht]
\centering
\includegraphics[width=\textwidth]{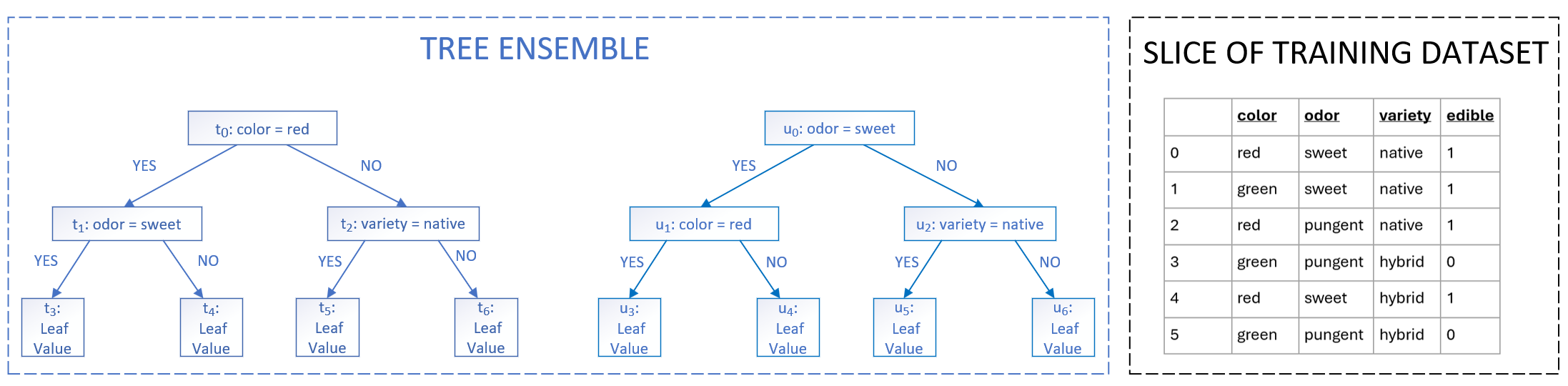} 
\vspace{-4mm}
\caption{An example of a tree ensemble with n = 2 trees each with depth d = 2 and a slice of data used to run TE2Rules. The tree ensemble uses features like color, odor, variety of a fruit to predict if it is edible. The positive class corresponds to edible = 1.}
\label{fig:illustration}
\vspace{-5mm}
\end{figure*}

\textbf{TE2Rules Algorithm}: TE2Rules takes a trained tree ensemble (TE) model with hundreds of decision trees and a slice of training data as input and gives a list of rules for positive class as output. Consider a tree ensemble containing $n$ trees, each with a maximum depth of $d$. These trees are made of internal nodes, and each node decides whether to move left (if feature$_i \leq t_i$) or right (if feature$_i > t_i$) based on a single feature. When we input a data point into this tree ensemble model, it traverses all $n$ trees, navigating internal nodes based on feature conditions. This journey involves passing through $d$ internal nodes + $1$ leaf node in each tree. Consequently, the data point traverses a total of $n(d + 1)$ tree nodes across the entire ensemble. It is important to note that any other data point meeting these $n(d+1)$ conditions would receive the same model prediction. Figure \ref{fig:illustration} shows a tree ensemble model with $n=2$ trees of depth $d=2$ and a slice of training data. The model predicts whether a fruit is edible based on 3 features: color, odor, variety. Here's a simplified overview of how TE2Rules operates on this tree ensemble.

\textbf{Step 1: Pre-Processing:} In this step, TE2Rules transforms each data point from the training data slice into an itemset of $n(d+1)$ tree nodes obtained from its journey through the tree ensemble, treating each node as a separate item. For example, in Figure \ref{fig:illustration}, consider the first data point in the training dataset, described by (color=red, odor=sweet, variety=native). This data point is transformed into the itemset \{$t_0, t_1, t_3, u_0, u_1, u_3$\}, encompassing all the nodes it traverses within the tree ensemble. Subsequently, the original training data is depicted as a collection of itemsets, with each data point having a corresponding itemset.

\textbf{Step 2: Apriori:} In this step, TE2Rules finds itemsets  that frequently appear alongside positive predictions among all the itemsets found in step 1. To achieve this, TE2Rules retains only the itemsets associated with positive model predictions and applies the Apriori algorithm to find sets of $k'$ nodes ($k' = 1, 2, 3, \ldots k$) that appear frequently. This step involves running multiple stages of Apriori and in stage $k'$, it identifies itemsets with $k'$ nodes. The Apriori algorithm is run with user defined parameters for $m$ = min\_support and $k$ = num\_stages.

\textbf{Step 3: Itemset-Rule:} In this step, TE2Rules converts each itemset found in step 2 into an itemset-rule. An itemset consists of a set of nodes. For a node to be visited by a data point, all decisions along the path from the root to that particular node (i.e. all its ancestors up to the root node) must be satisfied. Each node is represented by the corresponding rule needed to reach it. For example, to reach node $t_3$, the data point must satisfy the rule ``color=red and odor=sweet". Similarly, for visiting node $t_1$, the data point must satisfy the rule ``color=red". The root node $t_0$ is represented by the empty rule, since all data points start their journey from the root node.

Likewise, each itemset can be expressed as a rule formed by combining (via disjunction) the rules for each of its constituent nodes. For example, an itemset with nodes \{$t_1, t_3, u_2$\} would be represented by ``((color=red) and (color=red and odor=sweet) and (odor=sweet))" = ``(color=red and odor=sweet)". In this way, each itemset found by Apriori is converted into an itemset-rule.

\textbf{Step 4: High Precision Rules:} In this step, TE2Rules transforms every itemset-rule identified in step 3 into the rule: ``If itemset-rule Then model prediction = positive" and retains them only if their precision exceeds a user defined threshold of min\_precision. The precision of a rule measures its correctness. In this particular context, precision denotes the fraction: count( data points satisfying itemset-rule and model prediction = positive) / count(data points satisfying itemset-rule). By default, TE2Rules uses a min\_precision threshold of 0.95. Step 3 identifies itemsets that occur frequently with positives. This step helps in finding itemsets that frequently occur with positives but infrequently with negatives. For example, in Figure \ref{fig:illustration}, the itemset \{$t_2, u_1$\} representing the rule ``color$\neq$red and odor=sweet" occurs only with positives and never with negatives. Hence, TE2Rules would find the rule ``If color$\neq$red and odor=sweet, then model prediction = positive" as a possible rule to explain some of the positives.

\begin{table*}[ht]
\begin{tabular}{|p{0.02\linewidth} |p{0.60\linewidth} | p{0.07\linewidth} p{0.07\linewidth}| p{0.07\linewidth} p{0.07\linewidth}|}
\hline
\multirow{2}{*}{} & \multicolumn{1}{c|}{\multirow{2}{*}{\textbf{Rule}}} & \multicolumn{2}{c|}{\textbf{Precision}}    & \multicolumn{2}{c|}{\textbf{Recall}}     \\ 
\cline{3-6} & \multicolumn{1}{c|}{}  & \multicolumn{1}{l|}{\textbf{Train}} & \textbf{Test} & \multicolumn{1}{l|}{\textbf{Train}} & \textbf{Test} \\ 
\hline
1. &  priors\_count $>$ 2.5 \& age $\leq$ 36.5& \multicolumn{1}{l|}{ 0.973} & {0.957} & \multicolumn{1}{l|}{0.580} & {0.560}     \\ 
\hline
2. & priors\_count $\leq$ 12.5 \& age $\leq$ 21.5 \& sex\_Female $\leq$ 0.5 \& days\_arrest $\leq$ 17.5  &  \multicolumn{1}{l|}{ 0.967} & {0.952}  & \multicolumn{1}{l|}{0.155} & {0.156} \\ 
\hline
3. & priors\_count $>$ 5.5 \& days\_arrest $>$ -1.5  &  \multicolumn{1}{l|}{ 0.962} & {0.947} & \multicolumn{1}{l|}{0.404} & {0.421}   \\
\hline
4. & priors\_count $>$ 1.5 \& priors\_count $\leq$ 15.5 \& age $\leq$ 28.5 &  \multicolumn{1}{l|}{ 0.975} & {0.981} & \multicolumn{1}{l|}{0.409} & {0.382} 
 \\
\hline
5. & charge\_Felony $>$ 0.5 \& priors\_count $\leq$ 1.5 \& age $\leq$ 22.5 \& sex\_Male $>$ 0.5 &  \multicolumn{1}{l|}{ 0.952} & {0.893} & \multicolumn{1}{l|}{0.109} & {0.139}  \\
\hline
6. & days\_arrest $>$ 0.5 &  \multicolumn{1}{l|}{1.000} & {0.929} & \multicolumn{1}{l|}{0.036 } & {0.026 } \\
\hline
7. & priors\_count $>$ 4.5 \& age $\leq$ 51.5  &  \multicolumn{1}{l|}{ 0.969} & {0.957}  & \multicolumn{1}{l|}{0.503} & {0.518} \\
\hline
8. & priors\_count $>$ 12.5  \& days\_arrest $>$ -8.5   &  \multicolumn{1}{l|}{1.000} & {0.974}  & \multicolumn{1}{l|}{0.124} & {0.145}   \\
\hline
9. & priors\_count $>$ 0.5 \& priors\_count $\leq$ 12.5 \& age $\leq$ 23.5 \& days\_arrest $>$ -4.5  &  \multicolumn{1}{l|}{ 0.968} & {0.917}   & \multicolumn{1}{l|}{0.161} & {0.178}   \\ 
\hline
10. & priors\_count $\leq$ 2.5 \& age $\leq$ 22.5  \& race\_African\_American $>$ 0.5 \& sex\_Female $\leq$ 0.5  &  \multicolumn{1}{l|}{ 0.958} & {0.984}  & \multicolumn{1}{l|}{0.124} & {0.117} \\
\hline

\end{tabular}
\vspace{-2mm}
\caption{Rules generated by TE2Rules on a XGBoost model with 50 trees, depth 3, trained on compas dataset}
\label{tab: compas_rules}
\vspace{-4mm}
\vspace{-2mm}
\end{table*}

\textbf{Step 5: Post-Processing:} In this step, TE2Rules selects a small number of rules among all the rules found in step 4 to explain the entire tree ensemble model. This process resembles solving a set cover problem, where each rule covers positives, acting as a set. A greedy algorithm is employed, starting with an empty rule list and successively selecting rules that cover the highest number of positives not already covered by the list. The chosen rule is added to the list, and the process is repeated until all positive instances are covered. 

It's important to note that multiple rules may explain the same data, and the greedy algorithm randomly selects rules in case of ties. Another approach involves having domain experts review all rules generated by TE2Rules, assigning weights based on their alignment with human decision-making. This introduces a weighted set cover problem and addressing it is beyond the scope of this work. In this work, we use the same weight for all rules, and the greedy algorithm is used to address the set cover problem.

TE2Rules uses three user-defined parameters: min\_support, num\_stages, and min\_precision. TE2Rules uses a default value of $0.95$ for min\_precision. TE2Rules uses a very conservative value of $0$ as the default value for min\_support. This forces TE2Rules to explore all possible rules with support greater than $0$ to explain the positive model predictions. Users can speed up TE2Rules by setting a higher value for min\_support, skipping rules with little support in the training data. However, this may result in a slightly reduced fidelity of the rule list mined by TE2Rules. Unless otherwise specified, TE2Rules utilizes these default parameter values in all our experiments. In the results section, we describe the trade-off between fidelity and runtime with varying num\_stages.

\textbf{Example}: Here is an example of rule list generated by TE2Rules from a tree ensemble (TE) model. This TE model is an XGBoost model, consisting of 50 trees with a depth of 3 and was trained on the \emph{compas} dataset. The \emph{compas} dataset contains outcomes from a commercial algorithm assessing the likelihood of a convicted criminal to reoffend. Positive label indicates a high likelihood of reoffending. The underlying XGBoost model achieved an AUC of 0.765 on training data and 0.724 on test data. TE2Rules was used to explain the predictions of the XGBoost model using 10\% of the training data. TE2Rules uses only the model predictions on this slice of training data and does not have access to ground truth labels. TE2Rules was run till stage 3.

TE2Rules identified a list of 10 rules that collectively achieved a fidelity of 0.973 on the training data slice. These rules successfully explained all positive predictions (minority class) with a fidelity of 1.00 on positive model predictions, while achieving a fidelity of 0.956 on the negative model predictions. The rules generalized well to the test data, achieving a fidelity of 0.949 on overall test data, 0.994 on positive predictions, and 0.915 on negative predictions in test data.

The rules found by TE2Rules are shown in Table \ref{tab: compas_rules}, along with precision and recall of each rules on training/test data. Precision represents the fraction of data points satisfying the rule that are also identified as positive by the model, while recall is the fraction of positive model predictions covered by the rule. Notably, each rule identified by TE2Rules maintains a precision above $0.95$. It is essential to highlight that while these rules are accurate in explaining the model predictions, they may not align with how humans would have arrived at the same decision. These rules only highlight how the model arrived at the decision. These $10$ rules were condensed from $316$ rules found by TE2Rules at the end of Step 4. These selected $10$ rules in Step 5 represent just one of the many possible representations of the model explanations. For alternate representations, human input from domain experts would be necessary to select a different set of rules from the 316 rules based on which ones align closely with human decision-making.

\begin{figure*}[ht]
\centering
\vspace{-6mm}
\includegraphics[width=\textwidth]{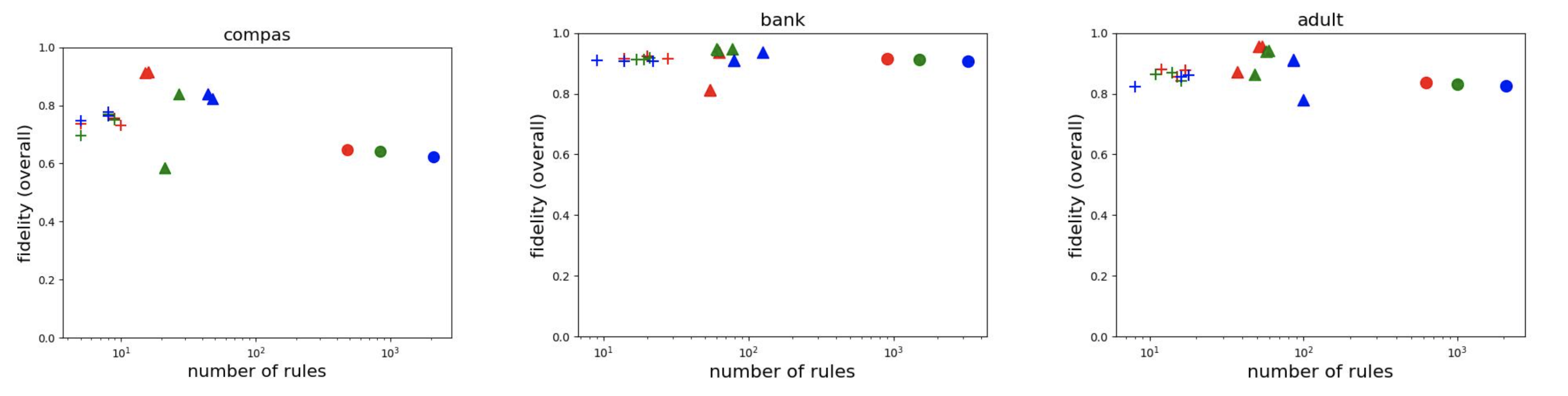} 
\vspace{-4mm}
\centering
\includegraphics[width=\textwidth]{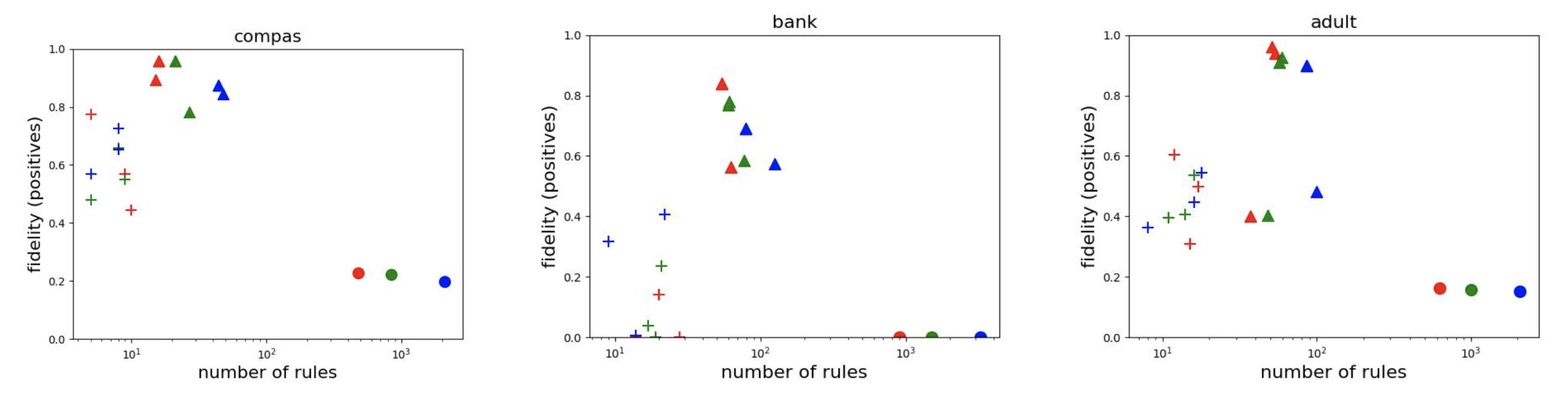} 
\vspace{-2mm}
\caption{Comparison of fidelity on test dataset (overall and on positives) versus number of rules mined for different explainers: TE2Rules ($\triangle$), inTrees ($\circ$), deFragTrees ($+$). All explainers are run on TE models with 100 (red), 200 (green), 500 (blue) trees of depth 5.}
\label{fig:fid}
\vspace{-2mm}
\end{figure*}

\section{Results}
\label{sec:results}
\textbf{Datasets}: We demonstrate the effectiveness of TE2Rules using 3 datasets from domains (like finance, legal sector) where transparency in decision making process is crucial: \emph{compas}, \emph{bank}, \emph{adult} \cite{compas, UCIdataset}. The \emph{compas} dataset consists of the results from a commercial algorithm used to assess a convicted criminal’s likelihood of reoffending. The \emph{bank} dataset consists of results from a marketing campaign by a banking institution on whether a client will subscribe to their term deposit. The \emph{adult} dataset consists of census data on whether a person has income over 50K\$. All these datasets contain demographic attributes of participants like age, gender, race, etc. 

\textbf{Baselines}: We compared TE2Rules with two popular baselines: inTrees and deFragTrees. 

inTrees goes through each node in the tree ensemble and extracts rules from each node using the decision path to reach the node from its respective root node. For each rule extracted from a node, it assigns the majority label from the support of the rule. Further, it selects a small set of high precision rules and presents it in a falling rule list format. 

deFragTrees identifies all possible rules that can be formed out of different node combinations from the tree ensemble. It then simplifies these rules by inducing a probability distribution over the rules and finding the simplest representation of this distribution. In this process, it finds a short falling rule list to represent the tree ensemble.

These algorithms operate at different ends of the spectrum. inTrees mines rules from individual nodes and completely discounts the effects of node combinations from multiple trees. deFragTrees mines rules by simplifying rules from all possible $n$ node combinations from multiple trees. TE2Rules provides a middle ground of exploring rules in stages of $k$ node combinations, with $k = 1,2,3, \ldots n$.


We report results of TE2Rules run with stages 1, 2 and 3. deFragTrees requires a parameter to specify the maximum number of rules to mine. We run deFragTrees with 1x, 5x and 10x times the number of rules mined by TE2Rules (with num\_stages = 3). Both the baselines (and TE2Rules) take the trained model and a sample of training data (10\%) as input to mine rules to explain the model. All explainers are run using the same sampled training dataset, trained model and evaluated using the same test dataset.  

\textbf{Implementation}: TE2Rules and deFragTrees are implemented in python while inTrees is implemented in R. We trained our xgboost models in python using scikit-learn and exported them in a format that can be ingested in R. Our implementation of TE2Rules with instructions to reproduce the results can be found here: \url{https://github.com/linkedin/TE2Rules}. All experiments were conducted on a 64-bit Ubuntu OS 20.04,  with an Intel Xeon 2.4 GHz CPU and 32 GB RAM.

\textbf{Models}: We trained gradient boosted tree ensemble (TE) models with 100, 200, 500 trees with depth 3, 5 for binary classification in python scikit-learn. In all our results, we use red, green, blue colors to denote TE models with 100, 200, 500 trees, respectively. We explain these models using inTrees, deFragTrees and TE2Rules. We report the number of extracted rules and time taken to extract the rules. We evaluate the performance of the rules using fidelity: accuracy of the rules with respect to the model predictions. We report the fidelity of the rules on the test data (overall fidelity) and on the portion of test data on which the model labels it as positive class (positive fidelity). In all these datasets, positive class happens to be the minority class.

\begin{figure*}[ht]
\centering
\vspace{-2mm}
\includegraphics[width=\textwidth]{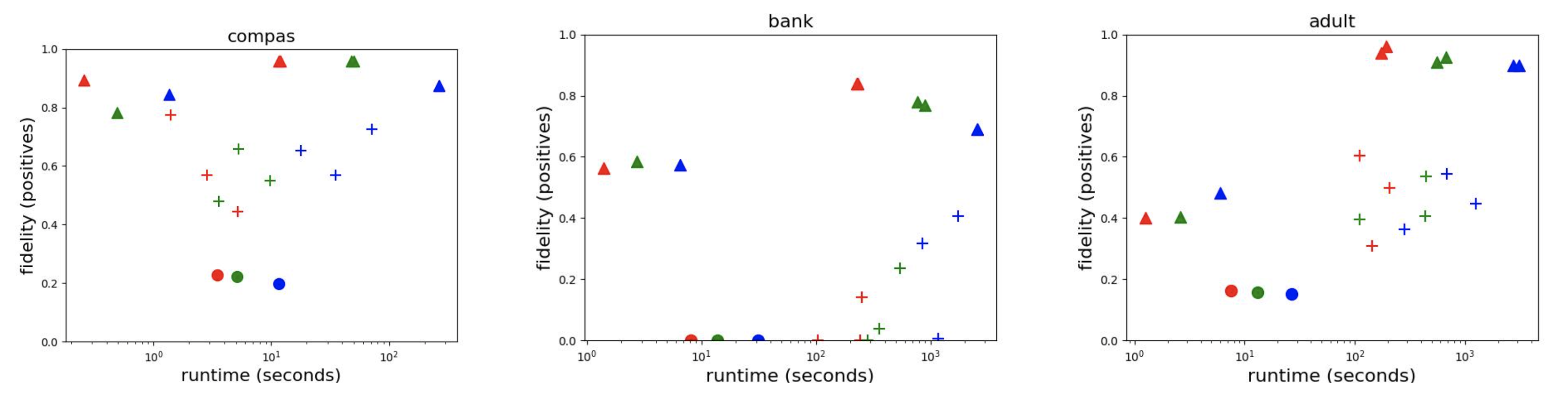} 
\vspace{-4mm}
\vspace{-2mm}
\caption{Comparison of fidelity (positives) on test data versus runtime for different explainers: TE2Rules ($\triangle$), inTrees ($\circ$), deFragTrees ($+$). All explainers are run on TE models with 100 (red), 200 (green), 500 (blue) trees of depth 5.}
\label{fig:time}
\vspace{-2mm}
\end{figure*}

\begin{figure*}[ht]
\centering
\vspace{-2mm}
\includegraphics[width=\textwidth]{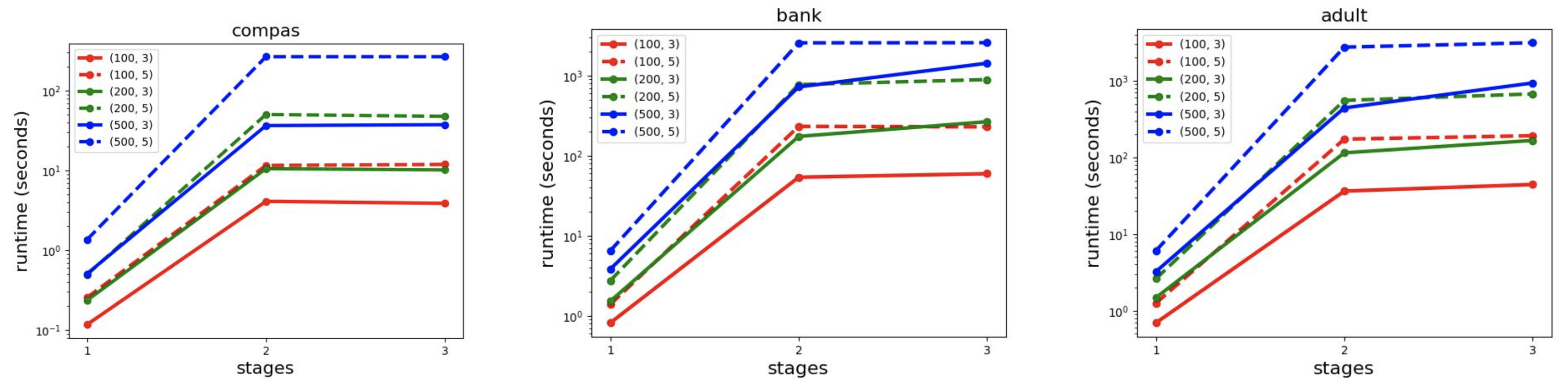} 
\vspace{-4mm}
\centering
\includegraphics[width=\textwidth]{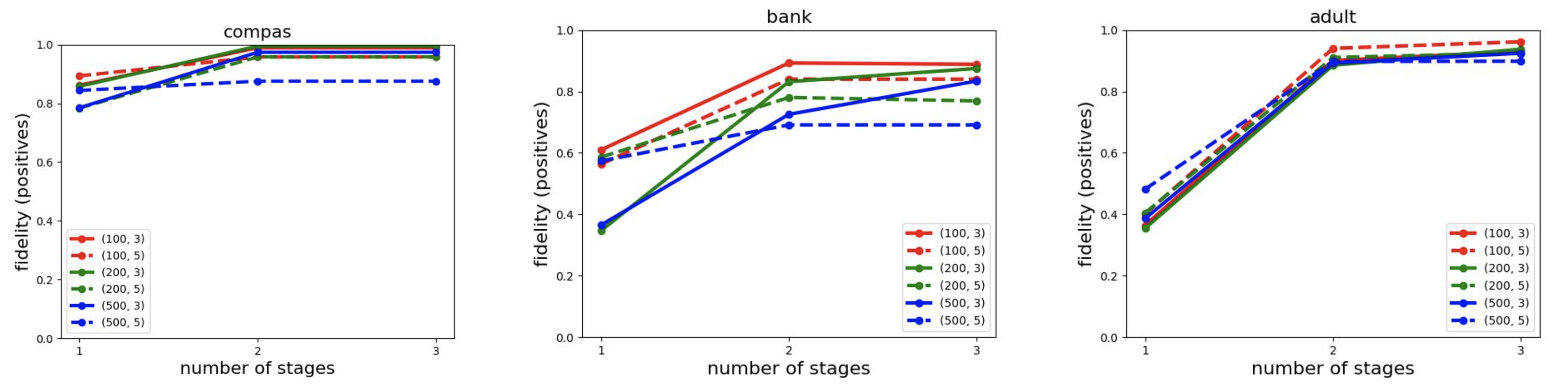} 
\vspace{-2mm}
\caption{Comparison of runtime and fidelity (positives) of TE2Rules with more number of stages. TE2Rules is run on TE models with 100 (red), 200 (green), 500 (blue) trees of depth 3 (solid line), 5 (dashed line).}
\label{fig:stages}
\vspace{-2mm}
\end{figure*}

\subsection{Performance: Fidelity}

Figure \ref{fig:fid} displays the fidelity of explainers on xgboost models with varying tree configurations (100, 200, 500 trees with depth 5). Each model is represented by a unique color, and comparisons between explainers ($+$: deFragTrees, $\triangle$: TE2Rules, $\circ$: inTrees) should be made within the same color. TE2Rules results are reported for stages 1, 2, and 3, each being an independent run involving stages 1-4 followed by post-processing (Step - 5). In each plot, three ``triangles" of the same color denote TE2Rules runs until stages 1, 2, and 3 for each xgboost model. Among triangles of the same color, the one with the lowest fidelity on positives corresponds to stage-1. While fidelity on positives generally improves with stages, stage-3 often provides minimal improvement, resulting in overlap with stage-2 triangles, indicating similar numbers of rules and fidelity on positives. Consequently, most plots within Figure \ref{fig:fid} typically show only 2 triangles.

In the top 3 plots of Figure \ref{fig:fid}, all explainers achieve very high fidelity on the overall test data. In the bottom 3 plots, baselines exhibit poor fidelity on the portion of the test set with positive model predictions, with deFragTrees outperforming inTrees. TE2Rules achieves higher fidelity than both inTrees and deFragTrees on positive model predictions.

TE2Rules achieves high fidelity on positive model predictions by combining rules from multiple trees, whereas inTrees only fetches rules from individual trees. DeFragTrees performs better by mining rules from the global model prediction boundaries. Although stage-1 of TE2Rules is closer to inTrees in principle, as it mines rules from individual nodes, TE2Rules outperforms inTrees. This is because the rule list mined by inTrees for the minority class (positives) is insufficient, mostly explaining the majority class (negatives) and resulting in poor performance in fidelity on positives. DeFragTrees faces a similar challenge. Despite mining 10 times as many rules as TE2Rules, deFragTrees struggles to explain the minority class effectively compared to TE2Rules stage-3. Therefore, TE2Rules proves to be more effective in explaining the minority class (positives) compared to inTrees and deFragTrees by mining rules from multiple node combinations within the tree ensemble.

\subsection{Number of Rules}

From Figure \ref{fig:fid}, we observe that TE2Rules mines more rules than inTrees but fewer rules than deFragTrees to explain xgboost models. Despite deFragTrees extracting a smaller number of rules compared to TE2Rules, it struggles to effectively explain the minority class (positives), exhibiting lower fidelity of positives compared to TE2Rules.

Among the various stages of TE2Rules, stage-1 (the triangle with lower fidelity on positives) often generates more rules than stages 2 and 3 (higher fidelity on positives). As illustrated in Figure \ref{fig:fid}, among the blue triangles in the bank dataset (bottom row, second plot), the one with higher positive fidelity (stage-2) has a lower number of rules than the one with lower positive fidelity (stage-1).

This is because the number of rules in the final output of TE2Rules consists of rules selected at the end of post-processing (Step-5). The number of rules at the end of Step 4 is always higher for TE2Rules run with more stages, as successive stages add more rules on top of each other. However, the post-processing step selects a small subset of rules from this pool. Therefore, the number of rules after post-processing can be lower for TE2Rules run until stage 2 (or 3) compared to the run until stage 1. This is particularly true because later stages may uncover more potent rules capable of explaining a greater number of positives that rules at the end of stage 1 simply cannot. Consequently, fewer such powerful rules are needed to account for all the positives. Thus, running TE2Rules for more stages can often result in a smaller number of rules with higher fidelity on positives.

\subsection{Scalability: Runtime}

Figure \ref{fig:time} illustrates the runtime and fidelity (on positives) of explainers for three XGBoost models: 100, 200, and 500 trees with a depth of 5. It also shows the runtime of different runs of TE2Rules run till stages 1, 2 and 3. In each color, triangle with lower fidelity on positives corresponds to the lower stages. In general, two clusters of $\triangle$ points corresponding to TE2Rules emerge: one on the left (lower runtime, lower fidelity on positives) and the other on the top right corner (higher runtime, higher fidelity on positives). The first cluster represents TE2Rules stage-1, while the second encompasses stages 2 and 3. Overlapping triangles for stages 2 and 3, as explained in the previous subsection, can occur.

In stage-1, TE2Rules achieves higher fidelity on positives compared to inTrees and with a shorter runtime. With stages 2 and 3, TE2Rules achieves even higher fidelity on positives, surpassing that of deFragTrees with runtimes that are comparable or slightly higher than  that of deFragTrees. Thus, TE2Rules demonstrates impressive performance in terms of fidelity on positives while maintaining runtime efficiency, making it a robust choice for explaining tree ensemble models.

\subsection{Fidelity-Runtime tradeoff}

Figure \ref{fig:stages} shows the run time and positive fidelity of TE2Rules with stages for 6 different xgboost models with 100, 200, 500 trees and depth 3, 5. We note that there is marginal improvement in fidelity on positives beyond stage-2. This shows that, most positives in the data can be explained within 2 stages of TE2Rules. So for all our use cases, running TE2Rules for 3 stages was sufficient. 

Similarly, the runtime of TE2Rules run till stage 3 is not very different from that of TE2Rules run till stage 2. A brute force search across node combinations would have meant exploring exponentially more nodes in every stage. But due to the smart way of generating stage-3 candidates from stage-2 using the Apriori Algorithm, very few (almost no) candidates with non-zero support are generated in stage-3. This reduces the runtime of stage-3 significantly. Since, most of the rules are mined in early stages, TE2Rules can be stopped early (2 to 3 stages) without loosing much fidelity on positives.  

%

\section{Conclusion}
\label{sec:conclusion}

We presented a novel approach, TE2Rules to explain a binary tree ensemble (TE) classifier using rules mined specially for a class of interest. We showed that our explainer is faithful (with high fidelity) to the model on both the overall test data and specifically on the minority class. It achieves such high performance in runtimes that are comparable to the state of the art baselines. Further, we show that stopping the algorithm in early stages can tradeoff runtime without loosing much fidelity on positives.

\bibliography{main}

\begin{thebibliography}{18}
\providecommand{\natexlab}[1]{#1}

\bibitem[{Agrawal, Srikant et~al.(1994)}]{apriori1994}
Agrawal, R.; Srikant, R.; et~al. 1994.
\newblock Fast algorithms for mining association rules.
\newblock In \emph{Proc. 20th int. conf. very large data bases, VLDB}, volume 1215, 487--499. Citeseer.

\bibitem[{Borgelt(2005)}]{fpgrowth2005}
Borgelt, C. 2005.
\newblock An Implementation of the FP-growth Algorithm.
\newblock In \emph{Proceedings of the 1st international workshop on open source data mining: frequent pattern mining implementations}, 1--5.

\bibitem[{Deng(2019)}]{deng2019interpreting}
Deng, H. 2019.
\newblock Interpreting tree ensembles with intrees.
\newblock \emph{International Journal of Data Science and Analytics}, 7(4): 277--287.

\bibitem[{Dua and Graff(2017)}]{UCIdataset}
Dua, D.; and Graff, C. 2017.
\newblock {UCI} Machine Learning Repository.

\bibitem[{Friedman and Popescu(2008)}]{rulefit2008}
Friedman, J.~H.; and Popescu, B.~E. 2008.
\newblock {Predictive learning via rule ensembles}.
\newblock \emph{The Annals of Applied Statistics}, 2(3): 916 -- 954.

\bibitem[{Gautier, Jafre, and Ndiaye(2020)}]{skoperules2021}
Gautier, R.; Jafre, G.; and Ndiaye, B. 2020.
\newblock scikit-learn-contrib/skope-rules.
\newblock \url{https://github.com/scikit-learn-contrib/skope-rules}.
\newblock V1.0.1.

\bibitem[{Grinsztajn, Oyallon, and Varoquaux(2022)}]{grinsztajn2022treebased}
Grinsztajn, L.; Oyallon, E.; and Varoquaux, G. 2022.
\newblock Why do tree-based models still outperform deep learning on tabular data?
\newblock arXiv:2207.08815.

\bibitem[{Guidotti et~al.(2018)Guidotti, Monreale, Ruggieri, Pedreschi, Turini, and Giannotti}]{lore}
Guidotti, R.; Monreale, A.; Ruggieri, S.; Pedreschi, D.; Turini, F.; and Giannotti, F. 2018.
\newblock Local Rule-Based Explanations of Black Box Decision Systems.
\newblock \emph{CoRR}, abs/1805.10820.

\bibitem[{Hara and Hayashi(2018)}]{defragtrees2018}
Hara, S.; and Hayashi, K. 2018.
\newblock Making tree ensembles interpretable: A bayesian model selection approach.
\newblock In \emph{International conference on artificial intelligence and statistics}, 77--85. PMLR.

\bibitem[{Lakkaraju, Bach, and Leskovec(2016)}]{lakkaraju2016interpretable}
Lakkaraju, H.; Bach, S.~H.; and Leskovec, J. 2016.
\newblock Interpretable decision sets: A joint framework for description and prediction.
\newblock In \emph{Proceedings of the 22nd ACM SIGKDD international conference on knowledge discovery and data mining}, 1675--1684.

\bibitem[{Larson et~al.(2016)Larson, Mattu, Kirchner, and Angwin}]{compas}
Larson, J.; Mattu, S.; Kirchner, L.; and Angwin, J. 2016.
\newblock How we analyzed the {COMPAS} recidivism algorithm. {ProPublica}.

\bibitem[{Meinshausen(2010)}]{node_harvest2010}
Meinshausen, N. 2010.
\newblock Node harvest.
\newblock \emph{The Annals of Applied Statistics}, 4(4).

\bibitem[{Qin et~al.(2021)Qin, Yan, Zhuang, Tay, Pasumarthi, Wang, Bendersky, and Najork}]{qin2021are}
Qin, Z.; Yan, L.; Zhuang, H.; Tay, Y.; Pasumarthi, R.~K.; Wang, X.; Bendersky, M.; and Najork, M. 2021.
\newblock Are Neural Rankers still Outperformed by Gradient Boosted Decision Trees?
\newblock In \emph{International Conference on Learning Representations}.

\bibitem[{Ribeiro, Singh, and Guestrin(2018)}]{anchors}
Ribeiro, M.~T.; Singh, S.; and Guestrin, C. 2018.
\newblock Anchors: High-Precision Model-Agnostic Explanations.
\newblock In \emph{AAAI}.

\bibitem[{Vidal, Pacheco, and Schiffer(2020)}]{vidal2020bornagain}
Vidal, T.; Pacheco, T.; and Schiffer, M. 2020.
\newblock Born-Again Tree Ensembles.
\newblock arXiv:2003.11132.

\bibitem[{Wang and Rudin(2015)}]{frl2015}
Wang, F.; and Rudin, C. 2015.
\newblock Falling rule lists.
\newblock In \emph{Artificial intelligence and statistics}, 1013--1022. PMLR.

\bibitem[{Wang et~al.(2017)Wang, Rudin, Doshi-Velez, Liu, Klampfl, and MacNeille}]{wang2017bayesian}
Wang, T.; Rudin, C.; Doshi-Velez, F.; Liu, Y.; Klampfl, E.; and MacNeille, P. 2017.
\newblock A bayesian framework for learning rule sets for interpretable classification.
\newblock \emph{The Journal of Machine Learning Research}, 18(1): 2357--2393.

\bibitem[{Yang, Rudin, and Seltzer(2017)}]{brl2017}
Yang, H.; Rudin, C.; and Seltzer, M. 2017.
\newblock Scalable Bayesian rule lists.
\newblock In \emph{International conference on machine learning}, 3921--3930. PMLR.

\end{thebibliography}



\end{document}